\documentclass{bmvc2k}

\title{EventFormer: AU Event Transformer for Facial Action Unit Event Detection}

\addauthor{Yingjie Chen}{chenyingjie@pku.edu.cn}{1}
\addauthor{Jiarui Zhang}{zjr954@pku.edu.cn}{1}
\addauthor{Tao Wang}{wangtao@pku.edu.cn}{1}
\addauthor{Yun Liang}{ericlyun@pku.edu.cn}{1}

\addinstitution{
 School of Computer Science \\
 Peking University\\
 Beijing, China
}

\runninghead{Chen~et al\bmvaOneDot}{EventFormer}

\def\etal{\emph{et al}\bmvaOneDot}

\usepackage{graphicx}
\usepackage{algorithm}
\usepackage{algorithmicx}
\usepackage{amssymb}
\usepackage{amsmath}
\usepackage{mathrsfs}
\usepackage{color}
\usepackage{multirow}
\usepackage{booktabs}
\usepackage{threeparttable}
\usepackage{dsfont}
\usepackage{wrapfig}
\usepackage{xspace}

\makeatletter
\DeclareRobustCommand\onedot{\futurelet\@let@token\@onedot}
\def\@onedot{\ifx\@let@token.\else.\null\fi\xspace}

\def\ie{\emph{i.e}\onedot} 
 
 \def\vs{\emph{vs}\onedot}
 
\makeatother

\begin{document}

\maketitle

\begin{abstract}
Facial action units (AUs) play an indispensable role in human emotion analysis.
We observe that although AU-based high-level emotion analysis is urgently needed by real-world applications, frame-level AU results provided by previous works cannot be directly used for such analysis. 
Moreover, as AUs are dynamic processes, the utilization of global temporal information is important but has been gravely ignored in the literature.
To this end, we propose EventFormer for AU event detection, which is the first work directly detecting AU events from a video sequence by viewing AU event detection as a multiple class-specific sets prediction problem.
Extensive experiments conducted on a commonly used AU benchmark dataset show the superiority of EventFormer under suitable metrics.
\end{abstract}

\section{Introduction}
Facial expression, as the most expressive emotional signal, plays an essential role in human emotion analysis. According to Facial Action Coding System (FACS)~\cite{FACS}, facial action units (AUs) refer to a set of facial muscle movements and are the basic components of almost all facial behaviors. The increasing need for user emotion analysis in application scenarios, such as online education and remote interview, leads to rapid growth in the field of AU analysis in recent years. 
With the prosperity of deep learning, two mainstream tasks of AU analysis, AU recognition~\cite{ciftci2017partially,GARN_2018_ACMMM, tirupattur2021modeling,chen2021cross,yang2020RENet,chen2021cafgraph} and AU intensity estimation~\cite{local_global_intensity, Fan_Lam_Li_2020_intensity, Song_2020_differentiable_intensity,fan2020joint,song2021dynamic}, have seen great improvements in recent years, both of which aim to estimate AU occurrence state or intensity for a given frame, \ie, frame-level AU results.

\begin{wrapfigure}[18]{r}{6.6cm}
\centering
    \includegraphics[scale=0.15]{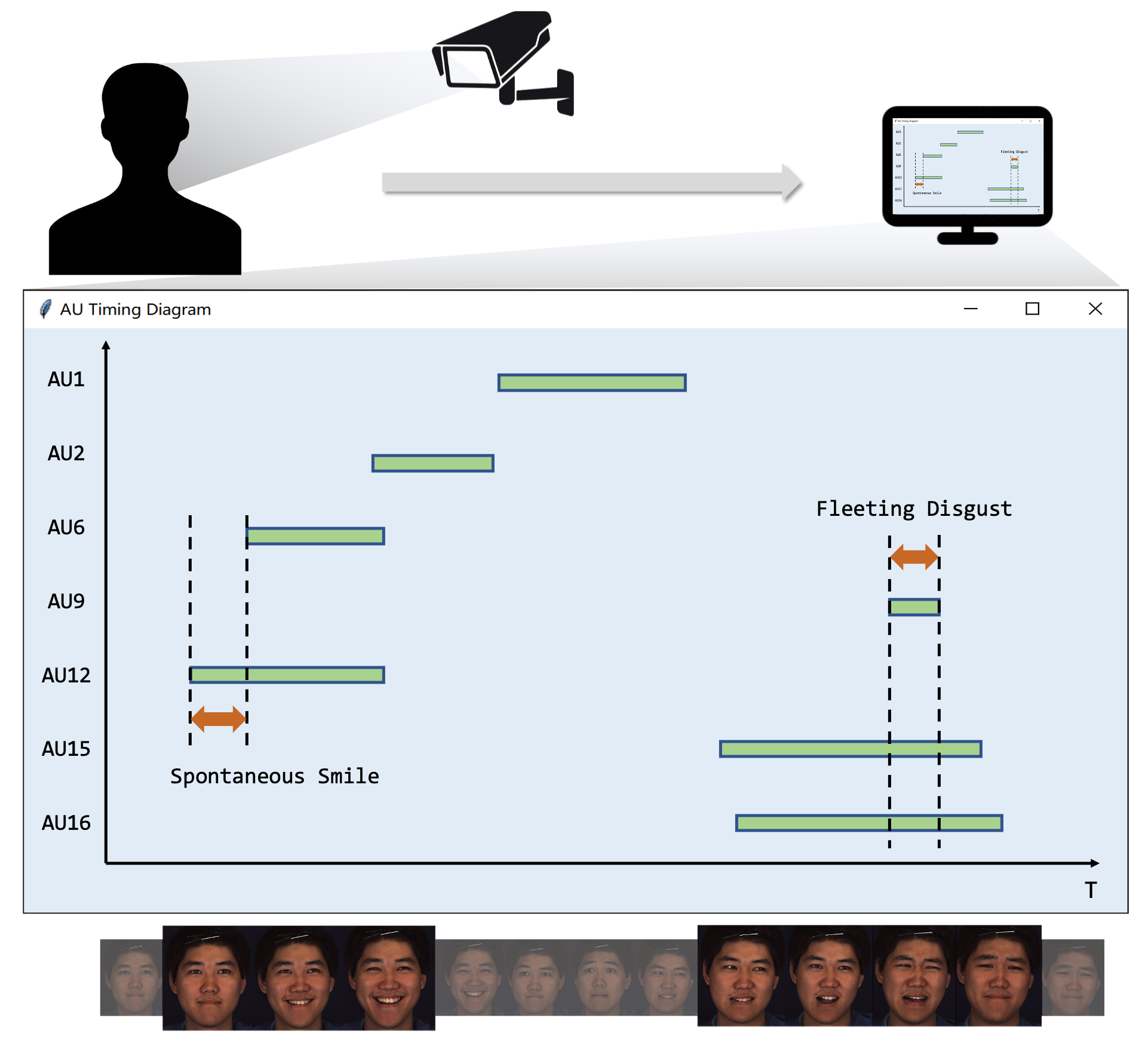}
    \caption{An illustration of the potential application scenario of direct sequence-level AU event detection. AU event provides overlapping as well as chronological information between different AUs, which is more suitable for further emotion analysis.}
\label{fig:motivation}
\end{wrapfigure}

However, when it comes to high-level emotion analysis, frame-level analysis results are not enough for the need of some real-world applications~\cite{schmidt2006movement, cohn2003timing}, due to the lack of various sequence-level information for further analysis, 
such as the occurrence frequencies, durations, and chronological order of AU events, each of which is a temporal segment containing one AU's complete temporal evolution, as shown in Fig.~\ref{fig:motivation}. 
For example, in public places such as airports, unnatural facial expressions or fleeting panics act as key information for the discrimination of a suspicious passenger. In this case, sequence-level AU event results are required to capture such abnormal emotions and based on which, warnings will be sent to officers for a further inspection of the person.
Furthermore, sequence-level AU event results are also important for the distinguishment between spontaneous and pretended facial expressions. 
For \emph{happiness}, the identification depends heavily on the overlapping situation of events labeled AU6 and AU12, in which case not only AU events' durations but also their chronological order matters.

Some works~\cite{chen2021aupro,ding2013facial} have made attempts to generate AU event results based on frame-level or unit-level (a fixed number of frames centered on the current one is regarded as a unit) AU results via a series of postprocessing steps, but they suffer from the lacking of global temporal information and are highly dependent on hyper-parameters for postprocessing.
To this end, we design an AU \textbf{Event} Trans\textbf{Former} (\textbf{EventFormer}) architecture to directly detect AU events from a video sequence by utilizing the benefit of global temporal information. 
EventFormer takes a video sequence as input and detects AU events for each AU class directly and simultaneously by viewing AU event detection as a \emph{multiple class-specific sets prediction problem}. 


\setlength\intextsep{0pt}

Specifically, first, a region-aware AU feature encoder is used for extracting fine-grained AU features as frame embedding. 
Then, an event transformer encoder-decoder module is used to generate event embeddings by learning global dependencies among frames in a video sequence.
Through self-attention mechanism, all frame embeddings are fed to transformer architecture simultaneously, and each frame can interact with others directly. In this way, a global view is maintained. After that, classification branch and regression branch are applied to each event embedding for event validity prediction and boundary regression, respectively.

Unlike methods such as~\cite{ROINet_2017_CVPR} modeling local temporal relations among several frames via RNN~\cite{LSTM} or methods such as~\cite{ding2013facial, 2010svm} using local temporal information by extracting unit-level features, our EventFormer takes the whole video sequence as input and model global temporal relations through the mechanism of transformer, which allows each frame to access all the other frames simultaneously.
And instead of generating AU events based on frame-level results through postprocessing which highly depends on manually selected hyper-parameters, EventFormer detects AU events in a direct way, and thus is able to alleviate discontinuous results.


The key contributions of our work are listed as:
\begin{itemize}
    \item To the best of our knowledge, this is the first work that directly detect AU events from a video sequence, which are more critical and practical for real-world applications.
    \item We propose EventFormer for AU event detection, taking advantage of the mechanism of transformer to maintain a temporal global view and alleviate discontinuous results.
    \item Extensive experiments conducted on a commonly used AU benchmark dataset, BP4D, show the superiority of our method under suitable metrics for AU event detection.
\end{itemize}

\section{Related Work}
In recent years, AU analysis tasks have drawn increasing attention as fundamental tasks in the field of affective computing. Conventional methods~\cite{simon2010action,2010svm} mainly design hand-crafted features as the input of a classifier for AU recognition. With the development of deep learning, methods~\cite{DRML_2016_CVPR, ROINet_2017_CVPR,fan2020joint,chen2022causal,chen2022pursuing} have raised the performance of AU analysis to a new height.
Since AUs are dynamic processes, methods such as~\cite{ROINet_2017_CVPR} employ LSTM to model local temporal relations, but longer the video sequence, weaker the relationships between temporally far apart frames. To obtain AU event results, Ding~\etal~\cite{ding2013facial} proposed a method that extracts unit-level features, predicts event-related scores and generates AU events via a series of postprocessing steps.
In contrast, we propose EventFromer to model global temporal relations among frames in a video sequence and detect AU events in a direct way.

Transformer~\cite{vaswani2017transformer} has attracted increasing research interest in computer vision tasks. Self-attention mechanism as the core of transformer allows the model to aggregate information from the whole input sequence with much less memory consumption and computing time compared to RNNs.
However, it is not until works~\cite{wu2020visualvit, carion2020detr} succeeded that the architecture has been proved effective and efficient in computer vision tasks. Our EventFormer makes full use of the mechanism to model global temporal information. 

\section{EventFormer for AU Event Detection}
\begin{figure}[!t]
    \centering
    \includegraphics[width=0.85\textwidth]{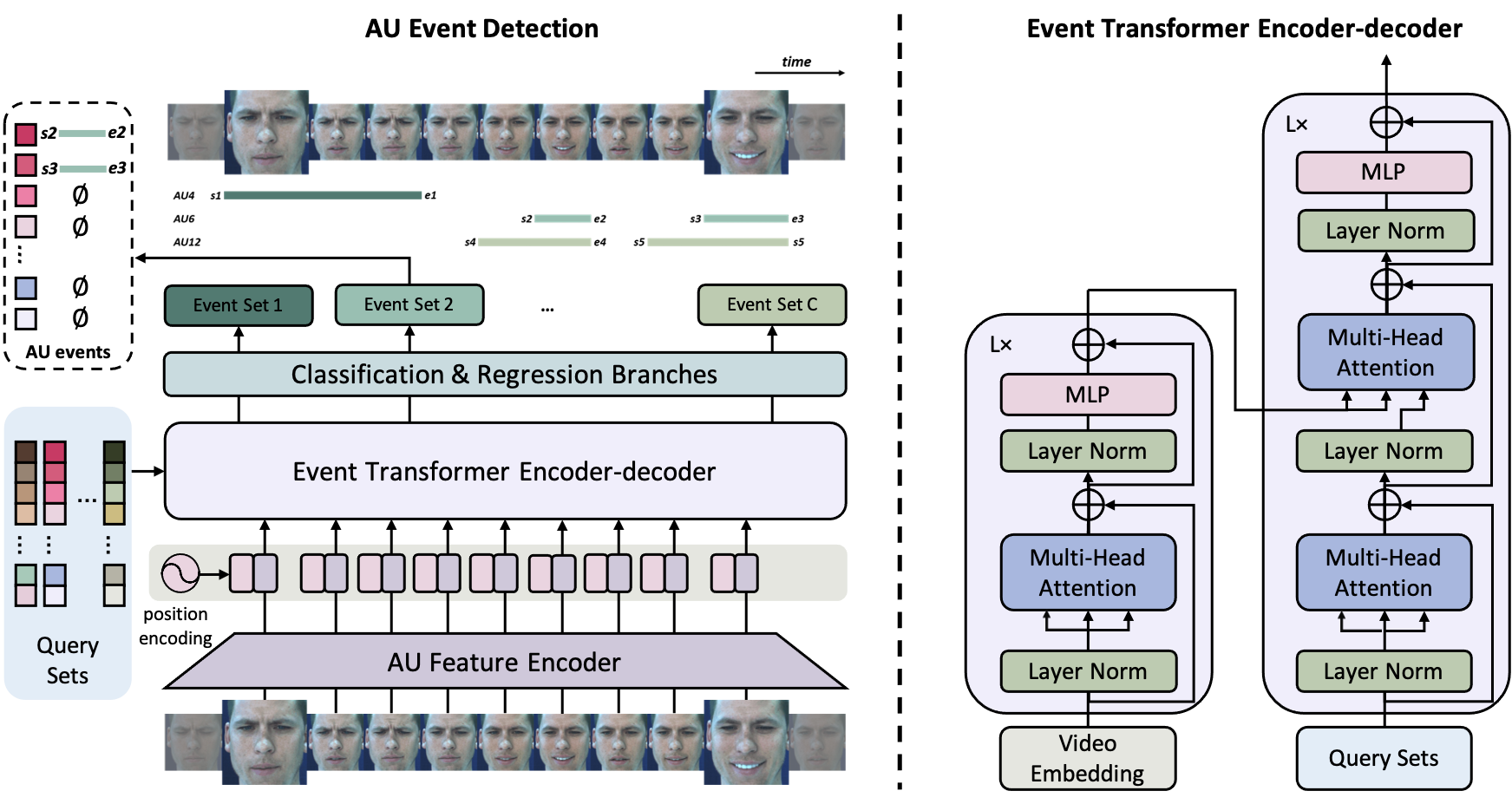}
    \caption{Architecture of EventFormer. EventFormer takes a video sequence as input, and AU feature encoder is first applied to the input to generate frame embeddings. Then position embeddings are concatenated to each frame embedding.
    Event transformer encoder models global relationships among frames via self-attention mechanism to enhance frame embeddings. After that, event transformer decoder takes encoder output and sets of queries, \ie~\emph{Query Sets}, as input, and outputs aggregated event embeddings for each query. Then the output event embeddings are passed to two branches to obtain the final \emph{Event Sets}.
    }
\label{fig:overview}
\end{figure}

\subsection{Problem Definition}
Given an input video sequence $\mathcal{I}=\{I_t\}^T_{t=1}$ with $T$ frames recording facial actions, where $I_t$ is the $t^{\rm th}$ frame in $\mathcal{I}$. The annotations of $\mathcal{I}$ are composed of a set of ground-truth AU events ${\Phi}_{\rm{g}}=\{{\phi}_i=(s^{\rm{g}}_i, e^{\rm{g}}_i, c^{\rm{g}}_i)|0\leq s^{\rm{g}}_i< e^{\rm{g}}_i\leq T$, $c^{\rm{g}}_i\in\{1, 2, \dots, C\}\}^M_{i=1}$, where $M$ is the number of ground-truth AU events in the video sequence $\mathcal{I}$, $C$ is the number of AU classes, and $s^{\rm{g}}_i$, $e^{\rm{g}}_i$, $c^{\rm{g}}_i$ are the start time, end time, and AU class label of AU events ${\phi}_i$, respectively.
AU event detection aims to detect a set of events ${\Phi}_{\rm p}=\{{\varphi}_i=(s^{\rm p}_i, e^{\rm p}_i, c^{\rm p}_i)|0\leq s^{\rm p}_i< e^{\rm p}_i\leq T$, $c^{\rm p}_i\in\{1, 2, \dots, C\}\}^N_{i=1}$ which match ${\Phi}_{\rm{g}}$ precisely and exhaustively.
During training, the set of ground-truth AU events ${\Phi}_{\rm{g}}$ is used as supervision for detected events ${\Phi}_{\rm p}$, and during inference, the ${\Phi}_{\rm p}$ can be simply filtered as results.

\vspace{-2mm}
\subsection{Multiple Class-specific Sets Prediction}
Due to AU co-occurrence relationships, AU events in different AU classes can have near-identical or exactly identical temporal boundaries.
Inspired by DETR~\cite{carion2020detr} which views object detection as a single set prediction problem, we view AU event detection as a multiple class-specific sets prediction problem. Instead of predicting a class-agnostic set with events of all classes, we predict multiple class-specific sets, each contains events for a specific AU class.

As noted above, AU event detection aims to detect class-specific sets of AU events, ${\Phi}_{\rm p}=\{{\varphi}_i=(s^{\rm p}_i, e^{\rm p}_i, c^{\rm p}_i)\}^N_{i=1}$, from $\mathcal{I}$.
If we bind AU class labels to events and search for a permutation of ${\Phi}_{\rm p}$ to match ${\Phi}_{\rm{g}}$ directly, some class mismatches caused by the multi-label property of AU event detection are hard to solve.
For example, events with identical temporal boundaries but different AU labels in ${\Phi}_{\rm{g}}$ can match with any permutation of the corresponding detected events in ${\Phi}_{\rm p}$ during training, which causes unstable training and makes the class labels hard to learn.
To alleviate the instability issue, we split ${\Phi}_{\rm{g}}$ into $C$ disjoint class-specific sets $\{{\Phi}^c_{\rm{g}}\}^{C}_{c=1}$ such that ${\Phi}_{\rm{g}}=\bigcup^{C}_{c=1}{\Phi}^c_{\rm{g}}$, where ${\Phi}^c_{\rm{g}}=\{{\phi}^c_i=(s^{\rm{g}}_i, e^{\rm{g}}_i, c^{\rm{g}}_i)|\forall {\phi}_i~\mathrm{ s.t.}~ c^{\rm{g}}_i=c\}^{N_c}_{i=1}$, and $N_c$ is the number of ground-truth events belonging to class $c$. 
In this way, the problem turns into predicting several class-specific sets ${\Phi}^c_{\rm p}$, (${\Phi}_{\rm p}=\bigcup^{C}_{c=1}{\Phi}^c_{\rm p}$), one for each AU class, \ie~a multiple class-specific sets prediction problem. 

For the implementation of EventFormer, assuming $N_0$ is a number larger than any $N_c$, we pad ${\Phi}^c_{\rm{g}}$ with $\varnothing$ (no event) to make the set $\Tilde{{\Phi}}^c_{\rm{g}}=\{\Tilde{\phi}^c_i=(s^{\rm{g}}_i, e^{\rm{g}}_i, v^{\rm{g}}_i)\}^{N_0}_{i=1}$ with a fixed number of events, where $v^{\rm{g}}_i\in \{0, 1\}$, $v^{\rm{g}}_i=0$ represents $\Tilde{\phi}^c_i$ is not a valid event, \ie~$\varnothing$ for padding, and $v^{\rm{g}}_i=1$ represents $\Tilde{\phi}^c_i$ is a valid event. We denote $\Tilde{{\Phi}}^c_{\rm p}=\{\Tilde{\varphi}^c_i=(s^{\rm p}_i, e^{\rm p}_i, v^{\rm p}_i)\}^{N_0}_{i=1}$ as the set of $N_0$ detected events for class $c$, called \emph{Event Set $c$}. 
EventFormer takes a video sequence $\mathcal{I}$ and a union of $C$ \emph{Query Sets} ${\Phi}_{\rm q}=\bigcup^{C}_{c=1}{\Phi}^c_{\rm q}$ as inputs, and outputs a union of $C$  corresponding \emph{Event Sets} $\Tilde{{\Phi}}_{\rm p}=\bigcup^{C}_{c=1}\Tilde{{\Phi}}^c_{\rm p}$.

\vspace{-2mm}
\subsection{EventFormer Architecture}
As shown in Fig.~\ref{fig:overview}, our EventFormer mainly consists of three parts.

\vspace{-4mm}
\paragraph{AU Feature Encoder}
Compared to coarse-grained body actions, AUs only cause subtle appearance changes on several local facial regions, which puts high demands on the discriminability of frame embeddings. 
Thus, we design a region-aware AU feature encoder to extract local features for each AU separately to preserve more detailed information.
Each frame $I_t\in \mathbb{R}^{3\times H\times W}$ in $\mathcal{I}$ is fed to a backbone network to extract $F^{\rm{global}}\in \mathbb{R}^{d\times H_0\times W_0}$ as global feature. And $C$ spatial attention layers~\cite{zhao2019pyramid} are applied to the global feature to extract local features $f^{\rm local}\in \mathbb{R}^{d}$ for each AU. Then, the concatenated local features $F^{\rm local}\in \mathbb{R}^{(d\times C)}$ are mapped to $E_t\in \mathbb{R}^{d_{\rm m}}$ via a linear layer as frame embedding for $I_t$.

\vspace{-4mm}
\paragraph{Event Transformer Encoder-decoder Module} 
We start from the original transformer encoder-decoder architecture~\cite{vaswani2017transformer} and design an event transformer encoder-decoder module specially for AU event detection.
Our event transformer encoder-decoder consists of $L$ encoder layers and $L$ decoder layers. To ensure a stable training period, LayerNorm~\cite{ba2016layer} is applied before Multi-head Attention and Multi-layer Perceptron, according to~\cite{xiong2020layer}. 
To maintain positional information in time dimension, positional encoding is employed to generate positional embeddings $\mathcal{P}=\{P_t\}^{T}_{t=1}\in \mathbb{R}^{T\times d_{\rm m}}$, corresponding to frame embeddings $\mathcal{E}=\{E_t\}^{T}_{t=1}\in \mathbb{R}^{T\times d_{\rm m}}$.
The transformer encoder takes the concatenated frame embeddings and positional embeddings, \ie~video embedding, as input and outputs refined frame embeddings by enabling interaction among frames via self-attention mechanism.
Event transformer decoder takes event queries $Q\in \mathbb{R}^{(CN_0)\times d_{\rm m}}$ as input, which can be regarded as the union of $C$ \emph{Query Sets}, ${\Phi}_{\rm q}=\bigcup^{C}_{c=1}{\Phi}^c_{\rm q}$, where ${\Phi}^c_{\rm q}=\{q^c_i\}^{N_0}_{i=1}$. 
Each query $q^c_i\in \mathbb{R}^{d_{\rm m}}$ is a learned positional embedding, which differs from each other. 
Queries first interact with each other through self-attention to alleviate event redundancy, and then interact with the encoder output, \ie~the refined frame embeddings as keys and values, to aggregate frame embeddings relative to each potential event as event embeddings $D\in \mathbb{R}^{(CN_0)\times d_{\rm m}}$. 

\vspace{-4mm}
\paragraph{Classification and Regression Branches}
The output event embeddings $D$ of the event transformer encoder-decoder module are further fed into classification branch and regression branch separately.
For each feature vector $d_i\in \mathbb{R}^{d_{\rm m}}$ in $D$ representing a potential event ${\varphi}_i$, the regression branch aims to estimate the start time $s_i$ and duration $l_i$ of the event, and $e_i=min(T, s_i+l_i)$.
The classification branch aims to estimate a one-hot vector $\hat{p}_i\in \mathbb{R}^2$ that denotes the probabilities of the value of $v_i$. 
We use a linear layer to output the classification probability $\hat{p}_i$ and two linear layers to regress $s_i$ and $l_i$.
After that, by reorganizing events in order, $\Tilde{{\Phi}}_{\rm p}=\bigcup^{C}_{c=1}\Tilde{{\Phi}}^c_{\rm p}$ is obtained.

\vspace{-3mm}
\section{Training and Inference of EventFormer}
To train EventFormer, a multiple class-specific sets matching cost is introduced for class-specific bipartite matching between each pair of $\Tilde{{\Phi}}^c_{\rm{g}}$ and $\Tilde{{\Phi}}^c_{\rm p}$. After the matching for each AU class, a multiple class-specific sets prediction loss can be computed for back-propagation.

\vspace{-2mm}
\subsection{Multiple Class-specific Sets Matching Cost}
Due to the disorder of events in one set, the loss function designed for multiple class-specific sets prediction should be invariant by a permutation of the detected events with identical class labels, \ie~in one \emph{Event Set}.
We apply a loss based on Hungarian algorithm~\cite{hungarian}, to find a bipartite matching between ground-truth events and detected ones for each class.

A permutation of $N_0$ elements $\sigma\in \Omega_{N_0}$ is searched by finding a bipartite matching between $\Tilde{{\Phi}}^c_{\rm{g}}$ and $\Tilde{\Phi}^c_{\rm p}$ for each class that minimizes the total matching cost, as shown in Eq.~\ref{eq:bipartite cost}:
\begin{equation}
    \hat{\sigma}={\arg\min}_{\sigma\in \Omega_{N_0}}\sum\nolimits^{N_0}_{i=1}\mathcal{L}_{\rm match}(\Tilde{\phi}^c_i, \Tilde{\varphi}^c_{\sigma(i)}),
\label{eq:bipartite cost}
\end{equation}
where $\mathcal{L}_{\rm match}(\Tilde{\phi}^c_i, \Tilde{\varphi}^c_{\sigma(i)f})$ is a pair-wise matching cost between ground-truth $\Tilde{\phi}^c_i$ and a detected event $\Tilde{\varphi}^c_{\sigma(i)}$ with matching index $\sigma(i)$. 
The loss function of matching is designed to minimize the distance between matched pairs and maximize the validity of matched events at the same time.
We define $\mathcal{L}_{\rm match}(\Tilde{\phi}^c_i, \Tilde{\varphi}^c_{\sigma(i)})$ as
\begin{equation}
    \mathds{1}_{\{v^{\rm{g}}_i=1\}}\left(\lambda_{\rm{bound}}\mathcal{L}_{\rm{bound}}(\Tilde{\phi}^c_i, \Tilde{\varphi}^c_{\sigma(i)}) -\lambda_{\rm{valid}}\hat{p}^c_{\sigma(i)}[v^{\rm{g}}_i]\right),
\label{eq:matching cost}
\end{equation}
where $\hat{p}^c_{\sigma(i)}\in \mathbb{R}^2$ denotes the probabilities of the value of $v^p_{\sigma(i)}$ indicating whether $\Tilde{\varphi}^c_{\sigma(i)}$ is a valid event, \ie the probabilities of $v^{\rm p}_{\sigma(i)}\in [0, 1]$, $\hat{p}^c_{\sigma(i)}[v^{\rm{g}}_i]$ denotes the probability of $v^{\rm p}_{\sigma(i)}=v^{\rm{g}}_i$, and $\lambda_{\rm{bound}}$ and $\lambda_{\rm{valid}}$ are for balancing. The boundary loss $\mathcal{L}_{\rm{bound}}$ measures the similarity between a pair of matched ground-truth event and detected event.
L1 loss measures the numerical difference of the regression results, and tIoU loss measures the overlapping area ratio of matched pairs. Both of them are used, considering pairs of ground-truth event and detected event may have a minor difference in terms of L1 but a huge difference in terms of tIoU.
Thus, a linear combination of tIoU loss (Eq.~\ref{eq:tiou loss}) and L1 loss is used as our boundary loss $\mathcal{L}_{\rm{bound}}$, as shown in Eq.~\ref{eq:boundary loss}.
\begin{equation}
\begin{aligned}
    & T_{\cap} = max(0, min(e_1, e_2) - max(s_1, s_2)), \\
    \mathcal{L}_{\rm tIoU}&((s_1, e_1), (s_2, e_2)) = \frac{T_{\cap}}{(e_1 - s_1) + (e_2 - s_2) + T_{\cap}}.
\label{eq:tiou loss}
\end{aligned}
\end{equation}
\begin{equation}
\begin{aligned}
    \mathcal{L}_{\rm{bound}}&(\Tilde{\phi}^c_i, \Tilde{\varphi}^c_{\sigma(i)})=\lambda_{\rm tIoU}\mathcal{L}_{\rm tIoU}((s^{\rm{g}}_i, e^{\rm{g}}_i), (s^{\rm p}_{\sigma(i)}, e^{\rm p}_{\sigma(i)})) \\
    & + \lambda_{\rm L1}(\Vert s^{\rm{g}}_i-s^{\rm p}_{\sigma(i)}\Vert+\Vert e^{\rm{g}}_i-e^{\rm p}_{\sigma(i)}\Vert),
\label{eq:boundary loss}
\end{aligned}
\end{equation}
where $\lambda_{\rm tIoU}$ and $\lambda_{\rm L1}$ are for balancing.

\vspace{-2mm}
\subsection{Multiple Class-specific Sets Prediction Loss}
After finding a bipartite matching minimizing the matching cost, the loss function can be computed. A combination of $\mathcal{L}_{\rm{bound}}$ and $\mathcal{L}_{\rm class}$ (Eq.~\ref{eq:classification loss}) forms the event detection loss $\mathcal{L}$, as shown in Eq.~\ref{eq:event detection loss}. 
\begin{equation}
    \mathcal{L}_{\rm class}(p^c_i,\hat{p}^c_{\sigma(i)}) = - \sum\nolimits_{v\in (0, 1)}p^c_i(v)\log(\hat{p}^c_{\sigma(i)}(v)).
\label{eq:classification loss}
\end{equation}
\begin{equation}
    \mathcal{L}=\sum\nolimits^{C}_{c=1}\sum\nolimits^{N_0}_{i=1}(\mathds{1}_{\{v^{\rm{g}}_i=1\}}\mathcal{L}_{\rm{bound}}(\Tilde{\phi}^c_i, \Tilde{\varphi}^c_{\sigma(i)}) + \lambda_{\rm class}\mathcal{L}_{\rm class}(p^c_i,\hat{p}^c_{\sigma(i)})),
\label{eq:event detection loss}
\end{equation}
where $p^c_i\in \mathbb{R}^2$ is the one-hot encoding of $v^g_i$ for $\Tilde{\phi}^c_i$, and $\lambda_{\rm class}$ is for balancing. It is worth mentioning that all the detected events are involved in the calculation of $\mathcal{L}_{\rm class}$, but only the matched events in \emph{Event Sets} are involved in the calculation of $\mathcal{L}_{\rm{bound}}$.

In the inference stage, bipartite matching is disabled.
By given a threshold $\tau$, we can simply filter out events with $\hat{p}_i$ lower than the threshold and preserve more valid events.
For each \emph{Event Set $c$}, each preserved event $\Tilde{\varphi}^c_i$ is assigned with an AU class label $c$ to form one final detected event ${\varphi}_i=(s^{\rm p}_i, e^{\rm p}_i, c^{\rm p}_i)$ with $c^{\rm p}_i=c$ in ${\Phi}_{\rm p}$. In this way, the set of final AU events ${\Phi}_{\rm p}$ can be easily obtained.

\section{Experiments}

\subsection{Experimental Settings}

\paragraph{Datasets \& Metrics}
Extensive experiments are conducted on a commonly used benchmark dataset, BP4D~\cite{BP4D}.
In BP4D, 328 videos of 41 participants are taken, including 23 women and 18 men. Each frame is annotated by certificated FACS coders with binary AU occurrence labels. We consider 12 emotion-related AUs on BP4D, including AU1, 2, 4, 6, 7, 10, 12, 14, 15, 17, 23 and 24.
To construct the training data, we use a sliding window with length $T$ to truncate all the videos into several equal length video sequences, and there is an overlap of $T\slash 2$. Based on binary AU occurrence labels, ground-truth AU events ${\Phi}_{\rm{g}}$ are obtained for each video sequence. 
Following the common protocol mentioned in~\cite{DRML_2016_CVPR}, subject-exclusive 3-fold cross-validation is conducted for all experiments.

Considering that AU event detection shares some similarities with temporal action detection~\cite{lin2020fast}, we select several suitable metrics for AU event detection drawing on those used in that task.
The goal of AU event detection task is to detect AU events with not only high precision but also acceptable recall.
Thus, we consider Mean Average Precision(mAP) and Average Recall with an average number of events (AR@AN) at different tIoU thresholds $\alpha$. $\alpha$ is set to $[0.3:0.1:0.7]$ for mAP and $[0.5:0.05:0.95]$ for AR@AN. We also report Area under the AR \vs~AN curve (AUC) for evaluation.

\vspace{-4mm}
\paragraph{Implementation Details}
All facial images are aligned and cropped according to facial landmarks and resized to $256\times 256$. 
RN50~\cite{resnet} without the last linear layer is used as backbone in AU feature encoder.
Empirically, we set local feature dimension $d=512$, $H_0=W_0=16$, embedding dimension $d_{\rm m}=256$, and the number of encoder/decoder layers $L$ is set to 6. Other hyper-parameters $\lambda_{\rm{bound}}$, $\lambda_{\rm{valid}}$, $\lambda_{tIoU}$, $\lambda_{L1}$ and $\lambda_{class}$ are set to 5, 1, 2, 5, 1, respectively. The number of queries $N_0$ in each \emph{Query Set} is set to 100, and $\tau$ is set to 0.5.
We train EventFormer with AdamW~\cite{adamw} optimizer setting transformer's learning rate to $10^{-4}$, AU feature encoder's learning rate to $10^{-5}$, and weight decay to $10^{-4}$. Batch size is set to 8 and the number of training epochs is set to 100. All models are trained on two NVIDIA Tesla V100 GPUs. 

\begin{table}[!t]
\scriptsize
\centering
\setlength{\tabcolsep}{0.7mm}{
\begin{tabular}{cc|ccccc|cccc}
    \toprule 
    Backbone & Scheme & mAP@0.3 & mAP@0.4 & mAP@0.5 & mAP@0.6 & mAP@0.7 & AR@10 & AR@50 & AR@100 & AUC \\
    \midrule
    \multirow{3}*{\shortstack{RN18}} 
    & Frame2Event~\cite{Zhao_2017_ICCV_SSN_TAG} & 10.03 & 8.97 & 8.07 & 7.17 & 6.22 & 3.83 & 24.15 & 60.15 & 27.34 \\
    & Unit2Event~\cite{chen2021aupro} & 22.09 & 19.52 & 16.71 & 14.53 & 12.54 & 48.22 & \textbf{73.83} & \textbf{74.51} & \textbf{68.17} \\
    & EventFormer & \textbf{33.82} & \textbf{29.09} & \textbf{24.34} & \textbf{20.29} & \textbf{16.39} & \textbf{51.46} & 62.27 & 68.34 & 60.06 \\
    \midrule
    \multirow{3}*{\shortstack{RN34}} 
    & Frame2Event~\cite{Zhao_2017_ICCV_SSN_TAG} & 14.33 & 12.20 & 10.53 & 8.80 & 7.23 & 3.93 & 18.58 & 41.37 & 20.09 \\
    & Unit2Event~\cite{chen2021aupro} & 24.04 & 21.18 & 18.15 & 15.84 & 13.77 & 49.09 & \textbf{73.79} & \textbf{75.85} & \textbf{68.49} \\
    & EventFormer & \textbf{36.92} & \textbf{32.06} & \textbf{26.87} & \textbf{22.37} & \textbf{18.18} & \textbf{52.83} & 64.95 & 69.62 & 62.12 \\
    \midrule
    \multirow{3}*{\shortstack{RN50}} 
    & Frame2Event~\cite{Zhao_2017_ICCV_SSN_TAG} & 16.56 & 14.32 & 12.50 & 10.63 & 8.88 & 4.16 & 18.94 & 40.36 & 20.30 \\
    & Unit2Event~\cite{chen2021aupro} & 24.40 & 22.01 & 19.36 & 17.03 & 14.67 & 50.24 & \textbf{73.46} & \textbf{76.15} & \textbf{68.24} \\
    & EventFormer & \textbf{41.41} & \textbf{35.79} & \textbf{30.10} & \textbf{25.00} & \textbf{20.32} & \textbf{53.59} & 66.72 & 72.31 & 63.76 \\
    \bottomrule
    \end{tabular}
}
\caption{Comparison among schemes on BP4D in terms of mAP@tIoU, AR@AN, AUC.}
\label{tab:comparison bp4d}
\end{table}

\vspace{-2mm}
\subsection{Comparison among Schemes}
We classify AU event detection schemes into three categories, which use frame-level \\ (Frame2Event), unit-level (Unit2Event), and video-level (Video2Event) results to detect AU events respectively.
To demonstrate the effectiveness of EventFormer, which follows the scheme of Video2Event, two methods following other schemes are selected for comparison. For a fair comparison, all methods including EventFormer apply the same AU feature encoder pre-trained using frame-level AU occurrence labels.

\vspace{-4mm}
\paragraph{Comparison to Frame2Event}
Frame2Event scheme collects frame-level AU results and converts them to AU events through postprocessing such as Temporal Actionness Grouping (TAG)~\cite{Zhao_2017_ICCV_SSN_TAG}, which involves two hyper-parameters, \emph{water level} $\gamma$ and \emph{union threshold} $\tau$. We use the same AU feature encoder and an MLP as classifier to obtain frame-level AU results.
Based on the predicted AU occurrence probabilities, candidate events are generated under several combinations of $\gamma$ and $\tau$. Specifically, we sample $\gamma$ and $\tau$ within the range of 0.5 and 0.95 with a step of 0.05. The confidence score of each candidate event $(s, e, c)$ is computed by averaging the probabilities of class $c$ within segment $(s, e)$. Soft-NMS~\cite{Bodla_2017_ICCV_softNMS} is employed to select $N_0$ events for each class from the candidate events.

As shown in Table~\ref{tab:comparison bp4d}, EventFormer outperforms Frame2Event method by a large margin in terms of mAP given any tIoU threshold, regardless of what backbone is used. Especially, EventFormer achieves a performance gain of $24.85\%$ in mAP@0.3 than Frame2Event method using RN50 as backbone, which shows the superiority of EventFormer. We also notice that Frame2Event method obtains a pretty low AR given a small AN, which is because the prediction jitters due to the lack of a global view make it hard to use a set of fixed hyper-parameters to balance the trade-off between AP and AR.
Such limitation reflects the necessity of maintaining a global view and detecting events directly.

\setlength\intextsep{0pt}
\begin{wraptable}[13]{r}{6.6cm}
\scriptsize
\centering
\begin{tabular}{c|c|c|c}
    \toprule
     Hyper-parameters & Values & mAP@0.5 & AUC \\
     \midrule
     \multirow{4}*{\shortstack{\#Queries in \emph{Query Set} \\ $N_0$}}
     & 10 & \textbf{31.78} & 34.77 \\
     & 50 & 31.33 & 45.77 \\
     & 100 & 30.03 & 62.32 \\
     & 200 & 25.04 & \textbf{63.68} \\
     \midrule
     \multirow{4}*{\shortstack{Embedding Dimension \\ $d_{\rm m}$}}
     & 128 & \textbf{30.22} & 59.65 \\
     & 256 & 30.03 & \textbf{62.32} \\
     & 512 & 24.82 & 61.20 \\
     & 1024 & 22.95 & 59.33 \\
     \midrule
     \multirow{4}*{\shortstack{\#Encoder/decoder layers \\ $L$}}
     & 3 & 29.39 & 58.67 \\
     & 4 & 29.58 & 59.51 \\
     & 5 & \textbf{30.31} & 62.05 \\
     & 6 & 30.03 & \textbf{62.32} \\
     \bottomrule
\end{tabular}
\caption{Sensitivity to hyper-parameters.}
\label{tab:hyper-param}
\end{wraptable}

\vspace{-4mm}
\paragraph{Comparison to Unit2Event}
We choose AUPro~\cite{chen2021aupro} on behalf of Unit2Event scheme, which extracts unit-level features and predicts event-related scores to generate AU events. AUPro estimates the start and end probabilities, $P_{\rm s}$ and $P_{\rm e}$, for each time position exhaustively, and generates an action completeness map $P_{\rm c}$ consisting of the completeness score for any event $(s, e)$, and the final confidence score for an event $(s,e)$ is calculated as $P_{\rm s}(s)\times P_{\rm e}(e)\times P_{\rm c}(s,e)$.
Since the method only predicts class-agnostic events, we simply make it generates $C$ sets of $P_{s}$, $P_{e}$, and $P_{c}$, one set for each class.
We adopt Soft-NMS to select $N_0$ events out of $T^2$ detected events for each class. 

As shown in Table~\ref{tab:comparison bp4d}, EventFormer outperforms Unit2Event method with any backbone in terms of mAP given any tIoU. Specifically, EventFormer achieves a performance gain of $17.01\%$ in mAP@0.3 than Unit2Event method with RN50 as backbone.
Since Unit2Event method generates events exhaustively, it is supposed to obtain better results in terms of AR.
Although EventFormer performs a little bit worse than Unit2Event method in terms of AR given a large AN, it outperforms it in AR@10 by $3.35\%$, which indicates that events generated by EventFormer are of better quality.

\vspace{-2mm}
\subsection{Sensitivity Analysis}
Table~\ref{tab:hyper-param} shows sensitivity analysis of hyper-parameters, including the number of queries in \emph{Query Set} $N_0$, embedding dimension $d_{\rm m}$ and the number of layers $L$ of encoder and decoder.
As the number of queries in \emph{Query Set} increases, mAP decreases while AUC increases, due to the trade-off between mAP and AR. We notice that the mAP does not decrease a lot from $N_0=10$ to $N_0=100$, and AUC increases much slower when $N_0>100$. Thus, we choose $N_0=100$ for EventFormer.
As for $d_{\rm m}$, AUC reaches its peak when we set $d_{\rm m}$ to 256, while at the same time, mAP is also around its best score.
As for the number of layers $L$, we notice that EventFormer achieves better performance with $L$ increasing. For the balance between computing complexity and model performance, we choose $L=6$ for EventFormer.

\vspace{-2mm}
\subsection{Class-agnostic Set~\vs~Class-specific Sets}
\begin{figure}[!t]
    \centering
    \includegraphics[scale=0.5]{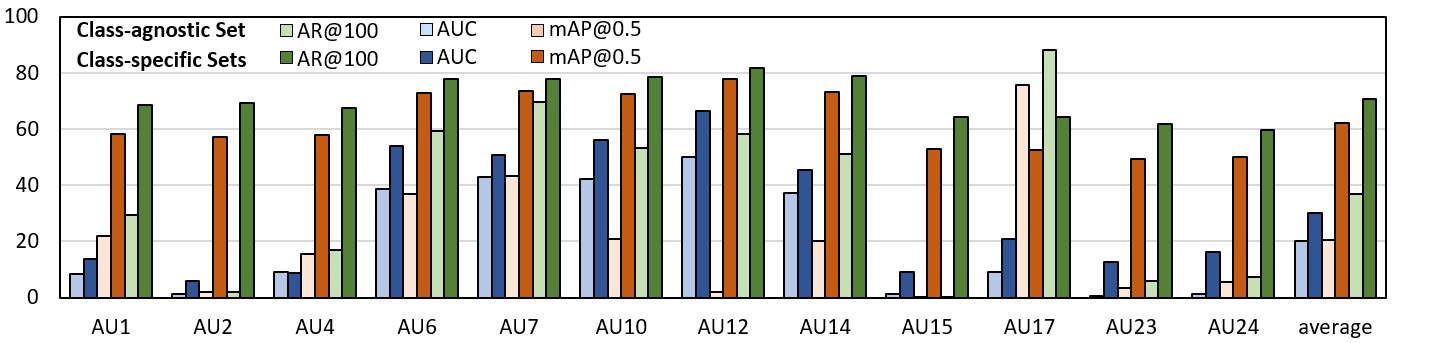}
    \caption{A comparison between multiple class-specific sets prediction and single class-agnostic set prediction.}
\label{fig:class-specific sets}
\end{figure}

To show the superiority of viewing AU event detection as a multiple class-specific sets prediction problem instead of a single class-agnostic set prediction problem, We implement a class-agnostic version of EventFormer for comparison, which generates events with AU class labels directly and applies bipartite matching once between the ground-truth events and detected ones of all classes.
From Fig.~\ref{fig:class-specific sets} we can see that the class-agnostic version obtains poor results on AU2, AU15, AU23, and AU24, of which the mAP@0.5 and AR@100 are near zero. The results variance among AU classes is huge for the class-agnostic version, while the class-specific version achieves relatively balanced results. We attribute the superiority to the binding between sets and AU classes, which is essential for stabilizing training and alleviating the variance of the results among AU classes caused by data imbalance problem.

\vspace{-4mm}
\subsection{Qualitative Results}

\paragraph{Visualization of Attention in EventFormer}

\setlength\intextsep{0pt}
\begin{wrapfigure}[9]{r}{6.6cm}
    \centering
    \includegraphics[scale=0.25]{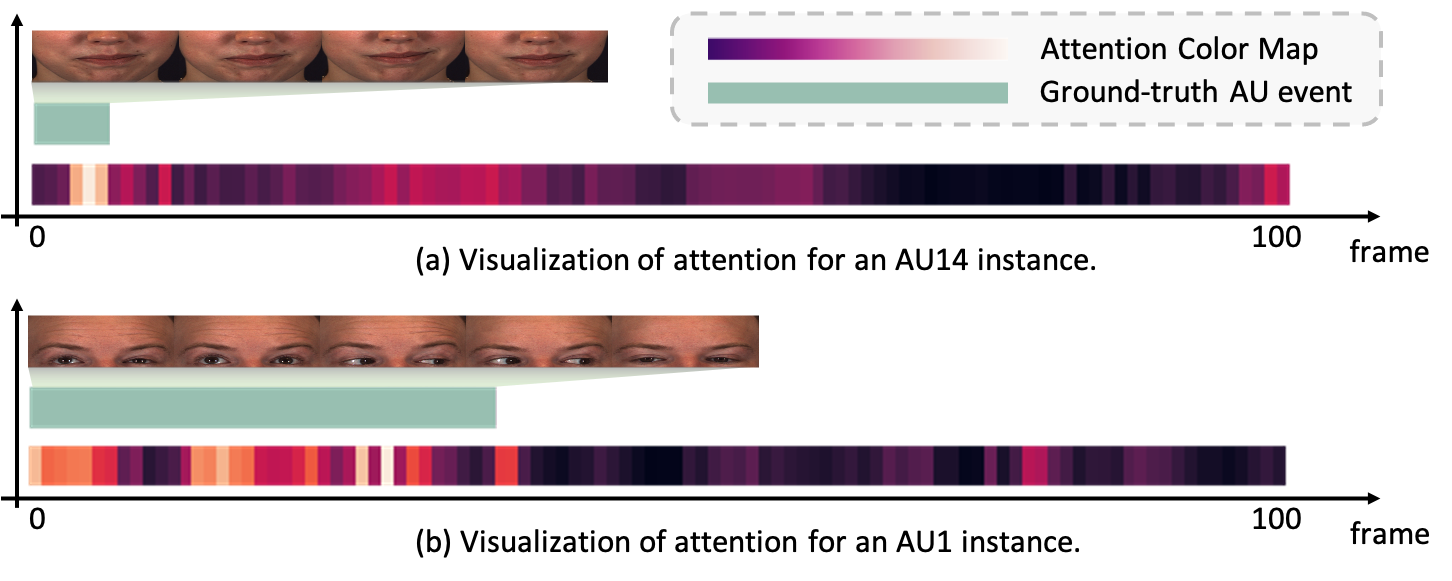}
    \caption{Visualization of attention.}
    \label{fig:attention}
\end{wrapfigure}

To better understand how the attention mechanism takes effect in EventFormer, we visualize the attention weights of the last layer of transformer decoder in Fig.~\ref{fig:attention}. The brightest parts of the attention show that the cross-attention of event transformer decoder tends to focus on the embeddings of frames where the states of AUs change. The results also show that EventFormer could capture subtle and transient appearance changes that occur in a very short duration ($\leq$5 frames) and detect an event successfully, as shown in Fig.~\ref{fig:attention}(a).

\setlength\intextsep{0pt}
\begin{wrapfigure}[11]{r}{6.6cm}
    \centering
    \includegraphics[scale=0.32]{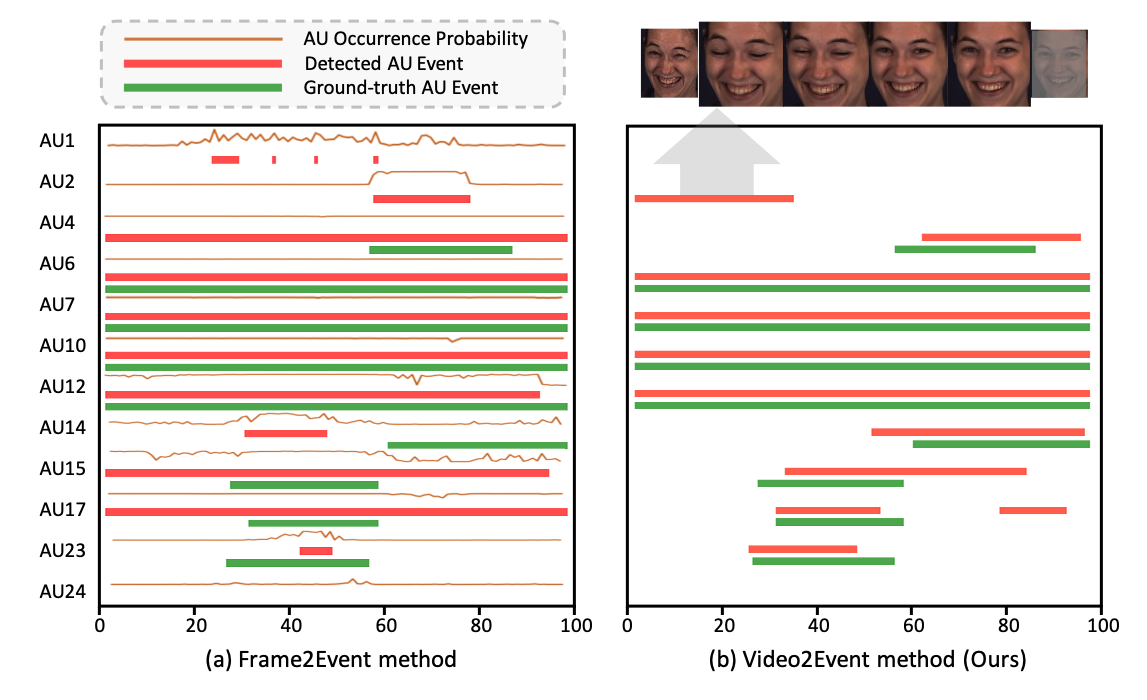}
    \caption{Visualization of detected AU events.}
\label{fig:timing_diagram}
\end{wrapfigure}

\vspace{-4mm}
\paragraph{Visualization of Detected AU Events}
Fig.~\ref{fig:timing_diagram} shows AU events detected by (a) Frame2Event method and (b) EventFormer.
AU events detected by EventFormer are of better quality, while AU events detected by Frame2Event method contains several false positive events with a very short duration.
There is a false positive event of AU2 (Outer Brow Raiser) in Fig~\ref{fig:timing_diagram}(b), and the visualized frames corresponding to this period show a process of the subject opening her eyes, during which wrinkles appeared above her eyebrow, misleading EventFormer to detect an AU2 event.

\vspace{-2mm}
\section{Conclusion}
This paper focuses on making full use of global temporal information to directly detect AU events from a whole video sequence, which are more practical and critical in some real-world application scenarios, such as financial anti-fraud. We propose EventFormer for AU event detection, which use the mechanism of transformer to model global temporal relationships among frames, and extensive experiments show the effectiveness of our EventFormer. 

\bibliography{egbib}

\begin{thebibliography}{36}
\providecommand{\natexlab}[1]{#1}
\providecommand{\url}[1]{\texttt{#1}}
\expandafter\ifx\csname urlstyle\endcsname\relax
  \providecommand{\doi}[1]{doi: #1}\else
  \providecommand{\doi}{doi: \begingroup \urlstyle{rm}\Url}\fi

\bibitem[Ba et~al.(2016)Ba, Kiros, and Hinton]{ba2016layer}
Jimmy~Lei Ba, Jamie~Ryan Kiros, and Geoffrey~E Hinton.
\newblock Layer normalization.
\newblock \emph{arXiv preprint arXiv:1607.06450}, 2016.

\bibitem[{Baltrušaitis} et~al.(2017){Baltrušaitis}, {Li}, and
  {Morency}]{local_global_intensity}
T.~{Baltrušaitis}, L.~{Li}, and L.~{Morency}.
\newblock Local-global ranking for facial expression intensity estimation.
\newblock In \emph{ACII}, pages 111--118, 2017.
\newblock \doi{10.1109/ACII.2017.8273587}.

\bibitem[Bodla et~al.(2017)Bodla, Singh, Chellappa, and
  Davis]{Bodla_2017_ICCV_softNMS}
Navaneeth Bodla, Bharat Singh, Rama Chellappa, and Larry~S. Davis.
\newblock Soft-nms -- improving object detection with one line of code.
\newblock In \emph{ICCV}, Oct 2017.

\bibitem[Carion et~al.(2020)Carion, Massa, Synnaeve, Usunier, Kirillov, and
  Zagoruyko]{carion2020detr}
Nicolas Carion, Francisco Massa, Gabriel Synnaeve, Nicolas Usunier, Alexander
  Kirillov, and Sergey Zagoruyko.
\newblock End-to-end object detection with transformers.
\newblock In \emph{ECCV}, pages 213--229. Springer, 2020.

\bibitem[Chen et~al.(2021{\natexlab{a}})Chen, Chen, Wang, Wang, and
  Liang]{chen2021cafgraph}
Yingjie Chen, Diqi Chen, Yizhou Wang, Tao Wang, and Yun Liang.
\newblock Cafgraph: Context-aware facial multi-graph representation for facial
  action unit recognition.
\newblock In \emph{Proceedings of the 29th ACM International Conference on
  Multimedia}, pages 1029--1037, 2021{\natexlab{a}}.

\bibitem[Chen et~al.(2021{\natexlab{b}})Chen, Wu, Wang, Wang, and
  Liang]{chen2021cross}
Yingjie Chen, Han Wu, Tao Wang, Yizhou Wang, and Yun Liang.
\newblock Cross-modal representation learning for lightweight and accurate
  facial action unit detection.
\newblock \emph{IEEE Robotics and Automation Letters}, 6\penalty0 (4):\penalty0
  7619--7626, 2021{\natexlab{b}}.

\bibitem[Chen et~al.(2021{\natexlab{c}})Chen, Zhang, Chen, Wang, Wang, and
  Liang]{chen2021aupro}
Yingjie Chen, Jiarui Zhang, Diqi Chen, Tao Wang, Yizhou Wang, and Yun Liang.
\newblock Aupro: Multi-label facial action unit proposal generation for
  sequence-level analysis.
\newblock In \emph{ICONIP}, pages 88--99. Springer, 2021{\natexlab{c}}.

\bibitem[Chen et~al.(2022{\natexlab{a}})Chen, Chen, Luo, Huang, Hua, Wang, and
  Liang]{chen2022pursuing}
Yingjie Chen, Chong Chen, Xiao Luo, Jianqiang Huang, Xian-Sheng Hua, Tao Wang,
  and Yun Liang.
\newblock Pursuing knowledge consistency: Supervised hierarchical contrastive
  learning for facial action unit recognition.
\newblock In \emph{Proceedings of the 30th ACM International Conference on
  Multimedia}, pages 111--119, 2022{\natexlab{a}}.

\bibitem[Chen et~al.(2022{\natexlab{b}})Chen, Chen, Wang, Wang, and
  Liang]{chen2022causal}
Yingjie Chen, Diqi Chen, Tao Wang, Yizhou Wang, and Yun Liang.
\newblock Causal intervention for subject-deconfounded facial action unit
  recognition.
\newblock In \emph{Proceedings of the AAAI Conference on Artificial
  Intelligence}, volume~36, pages 374--382, 2022{\natexlab{b}}.

\bibitem[Ciftci et~al.(2017)Ciftci, Zhang, and Tin]{ciftci2017partially}
UmurAybars Ciftci, Xing Zhang, and Lijun Tin.
\newblock Partially occluded facial action recognition and interaction in
  virtual reality applications.
\newblock In \emph{ICME}, pages 715--720. IEEE, 2017.

\bibitem[Cohn and Schmidt(2003)]{cohn2003timing}
Jeffrey~F Cohn and Karen Schmidt.
\newblock The timing of facial motion in posed and spontaneous smiles.
\newblock In \emph{Active Media Technology}, pages 57--69. World Scientific,
  2003.

\bibitem[Ding et~al.(2013)Ding, Chu, De~la Torre, Cohn, and
  Wang]{ding2013facial}
Xiaoyu Ding, Wen-Sheng Chu, Fernando De~la Torre, Jeffery~F Cohn, and Qiao
  Wang.
\newblock Facial action unit event detection by cascade of tasks.
\newblock In \emph{ICCV}, pages 2400--2407, 2013.

\bibitem[Ekman and Friesen(1978)]{FACS}
P.~Ekman and W.~Friesen.
\newblock \emph{Facial action coding system: A technique for the measurement of
  facial movement}.
\newblock 1978.

\bibitem[Fan et~al.(2020{\natexlab{a}})Fan, Shen, Cheng, and
  Tian]{fan2020joint}
Yachun Fan, Jie Shen, Housen Cheng, and Feng Tian.
\newblock Joint facial action unit intensity prediction and region
  localisation.
\newblock In \emph{2020 IEEE International Conference on Multimedia and Expo
  (ICME)}, pages 1--6. IEEE, 2020{\natexlab{a}}.

\bibitem[Fan et~al.(2020{\natexlab{b}})Fan, Lam, and
  Li]{Fan_Lam_Li_2020_intensity}
Yingruo Fan, Jacqueline Lam, and Victor Li.
\newblock Facial action unit intensity estimation via semantic correspondence
  learning with dynamic graph convolution.
\newblock In \emph{AAAI}, volume~34, pages 12701--12708, 2020{\natexlab{b}}.

\bibitem[He et~al.(2016)He, Zhang, Ren, and Sun]{resnet}
Kaiming He, Xiangyu Zhang, Shaoqing Ren, and Jian Sun.
\newblock Deep residual learning for image recognition.
\newblock In \emph{CVPR}, 2016.

\bibitem[Hochreiter and Schmidhuber(1996)]{LSTM}
Sepp Hochreiter and J{\"{u}}rgen Schmidhuber.
\newblock {LSTM} can solve hard long time lag problems.
\newblock In \emph{NIPS}, 1996.

\bibitem[Kuhn(1955)]{hungarian}
H.~W. Kuhn.
\newblock The hungarian method for the assignment problem.
\newblock \emph{Naval Research Logistics Quarterly}, 2\penalty0
  (1‐2):\penalty0 83--97, 1955.

\bibitem[Li et~al.(2017)Li, Abtahi, and Zhu]{ROINet_2017_CVPR}
Wei Li, Farnaz Abtahi, and Zhigang Zhu.
\newblock Action unit detection with region adaptation, multi-labeling learning
  and optimal temporal fusing.
\newblock In \emph{CVPR}, 2017.

\bibitem[Lin et~al.(2020)Lin, Li, Wang, Tai, Luo, Cui, Wang, Li, Huang, and
  Ji]{lin2020fast}
Chuming Lin, Jian Li, Yabiao Wang, Ying Tai, Donghao Luo, Zhipeng Cui, Chengjie
  Wang, Jilin Li, Feiyue Huang, and Rongrong Ji.
\newblock Fast learning of temporal action proposal via dense boundary
  generator.
\newblock In \emph{AAAI}, volume~34, pages 11499--11506, 2020.

\bibitem[Loshchilov and Hutter(2017)]{adamw}
Ilya Loshchilov and Frank Hutter.
\newblock Decoupled weight decay regularization.
\newblock \emph{arXiv preprint arXiv:1711.05101}, 2017.

\bibitem[Schmidt et~al.(2006)Schmidt, Ambadar, Cohn, and
  Reed]{schmidt2006movement}
Karen~L Schmidt, Zara Ambadar, Jeffrey~F Cohn, and L~Ian Reed.
\newblock Movement differences between deliberate and spontaneous facial
  expressions: Zygomaticus major action in smiling.
\newblock \emph{Journal of nonverbal behavior}, 30\penalty0 (1):\penalty0
  37--52, 2006.

\bibitem[{Simon} et~al.(2010){Simon}, {Nguyen}, {De La Torre}, and
  {Cohn}]{2010svm}
T.~{Simon}, M.~H. {Nguyen}, F.~{De La Torre}, and J.~F. {Cohn}.
\newblock Action unit detection with segment-based svms.
\newblock In \emph{2010 IEEE Computer Society Conference on Computer Vision and
  Pattern Recognition}, pages 2737--2744, 2010.
\newblock \doi{10.1109/CVPR.2010.5539998}.

\bibitem[Simon et~al.(2010)Simon, Nguyen, De~La~Torre, and
  Cohn]{simon2010action}
Tomas Simon, Minh~Hoai Nguyen, Fernando De~La~Torre, and Jeffrey~F Cohn.
\newblock Action unit detection with segment-based svms.
\newblock In \emph{CVPR}, pages 2737--2744. IEEE, 2010.

\bibitem[Song et~al.(2021)Song, Cui, Wang, Zheng, and Ji]{song2021dynamic}
Tengfei Song, Zijun Cui, Yuru Wang, Wenming Zheng, and Qiang Ji.
\newblock Dynamic probabilistic graph convolution for facial action unit
  intensity estimation.
\newblock In \emph{Proceedings of the IEEE/CVF Conference on Computer Vision
  and Pattern Recognition}, pages 4845--4854, 2021.

\bibitem[Song et~al.(2020)Song, Shi, Feng, Song, Lin, Lin, Fan, and
  Yuan]{Song_2020_differentiable_intensity}
Xinhui Song, Tianyang Shi, Zunlei Feng, Mingli Song, Jackie Lin, Chuanjie Lin,
  Changjie Fan, and Yi~Yuan.
\newblock Unsupervised learning facial parameter regressor for action unit
  intensity estimation via differentiable renderer.
\newblock In \emph{ACM MM}, pages 2842--2851, 2020.

\bibitem[Tirupattur et~al.(2021)Tirupattur, Duarte, Rawat, and
  Shah]{tirupattur2021modeling}
Praveen Tirupattur, Kevin Duarte, Yogesh~S Rawat, and Mubarak Shah.
\newblock Modeling multi-label action dependencies for temporal action
  localization.
\newblock In \emph{CVPR}, pages 1460--1470, 2021.

\bibitem[Vaswani et~al.(2017)Vaswani, Shazeer, Parmar, Uszkoreit, Jones, Gomez,
  Kaiser, and Polosukhin]{vaswani2017transformer}
Ashish Vaswani, Noam Shazeer, Niki Parmar, Jakob Uszkoreit, Llion Jones,
  Aidan~N Gomez, Lukasz Kaiser, and Illia Polosukhin.
\newblock Attention is all you need.
\newblock In \emph{NIPS}, 2017.

\bibitem[Wang and Wang(2018)]{GARN_2018_ACMMM}
Can Wang and Shangfei Wang.
\newblock Personalized multiple facial action unit recognition through
  generative adversarial recognition network.
\newblock In \emph{ACM MM}, pages 302--310, 2018.

\bibitem[Wu et~al.(2020)Wu, Xu, Dai, Wan, Zhang, Yan, Tomizuka, Gonzalez,
  Keutzer, and Vajda]{wu2020visualvit}
Bichen Wu, Chenfeng Xu, Xiaoliang Dai, Alvin Wan, Peizhao Zhang, Zhicheng Yan,
  Masayoshi Tomizuka, Joseph Gonzalez, Kurt Keutzer, and Peter Vajda.
\newblock Visual transformers: Token-based image representation and processing
  for computer vision.
\newblock \emph{arXiv preprint arXiv:2006.03677}, 2020.

\bibitem[Xiong et~al.(2020)Xiong, Yang, He, Zheng, Zheng, Xing, Zhang, Lan,
  Wang, and Liu]{xiong2020layer}
Ruibin Xiong, Yunchang Yang, Di~He, Kai Zheng, Shuxin Zheng, Chen Xing,
  Huishuai Zhang, Yanyan Lan, Liwei Wang, and Tieyan Liu.
\newblock On layer normalization in the transformer architecture.
\newblock In \emph{ICML}, pages 10524--10533. PMLR, 2020.

\bibitem[Yang and Yin(2020)]{yang2020RENet}
Huiyuan Yang and Lijun Yin.
\newblock Re-net: A relation embedded deep model for au occurrence and
  intensity estimation.
\newblock In \emph{Proceedings of the Asian Conference on Computer Vision},
  2020.

\bibitem[Zhang et~al.(2014)Zhang, Yin, Cohn, Canavan, Reale, Horowitz, Liu, and
  Girard]{BP4D}
Xing Zhang, Lijun Yin, Jeffrey~F Cohn, Shaun Canavan, Michael Reale, Andy
  Horowitz, Peng Liu, and Jeffrey~M Girard.
\newblock Bp4d-spontaneous: A high-resolution spontaneous 3d dynamic facial
  expression database.
\newblock \emph{Image and Vision Computing}, 32\penalty0 (10):\penalty0
  692--706, 2014.

\bibitem[Zhao et~al.(2016)Zhao, Chu, and Zhang]{DRML_2016_CVPR}
Kaili Zhao, Wen{-}Sheng Chu, and Honggang Zhang.
\newblock Deep region and multi-label learning for facial action unit
  detection.
\newblock In \emph{CVPR}, 2016.

\bibitem[Zhao and Wu(2019)]{zhao2019pyramid}
Ting Zhao and Xiangqian Wu.
\newblock Pyramid feature attention network for saliency detection.
\newblock In \emph{CVPR}, pages 3085--3094, 2019.

\bibitem[Zhao et~al.(2017)Zhao, Xiong, Wang, Wu, Tang, and
  Lin]{Zhao_2017_ICCV_SSN_TAG}
Yue Zhao, Yuanjun Xiong, Limin Wang, Zhirong Wu, Xiaoou Tang, and Dahua Lin.
\newblock Temporal action detection with structured segment networks.
\newblock In \emph{ICCV}, Oct 2017.

\end{thebibliography}
\end{document}